\PassOptionsToPackage{pagebackref=true,hidelinks}{hyperref}
\documentclass[final]{style/opt2024}

\usepackage[utf8]{inputenc}
\usepackage[T1]{fontenc}

\usepackage[lines=3]{lettrine}
\usepackage{style/Zallman}
\usepackage[nopatch=footnote]{microtype}
\usepackage{fontawesome}
\usepackage{dsfont}
\usepackage{lmodern}
\usepackage{anyfontsize}
\usepackage{verbatim}
\usepackage{pifont}

\usepackage{quoting}
\usepackage{xcolor}

\usepackage{url}

\usepackage{tcolorbox}
\usepackage{booktabs,array}
\usepackage{makecell}
\usepackage{multirow}
\usepackage{tikz}
\usetikzlibrary{shapes.geometric,positioning}
\usepackage{pgfornament}

\usepackage[frozencache,cachedir=minted-output]{minted}

\usepackage{amsfonts,amssymb,amsmath}
\usepackage[noabbrev,nameinlink,capitalise,compress]{cleveref}
\usepackage{mathtools}
\usepackage{bm}
\usepackage{mathabx}
\usepackage{thmtools,thm-restate}
\usepackage{resizegather}
\usepackage{nicefrac}
\usepackage{xfrac}

\usepackage{enumitem}
\usepackage{setspace}
\usepackage{regexpatch}

\DeclareMathOperator*{\argmax}{arg\,max}
\DeclareMathOperator*{\argmin}{arg\,min}

\newcommand{\defeq}{\vcentcolon=}
\newcommand{\el}{\mathcal{L}}

\newcommand{\rms}{\mathrm{RMS}}
\newcommand{\out}{\mathrm{out}}
\newcommand{\inn}{\mathrm{in}}

\renewcommand{\phi}{\varphi}

\newcommand{\half}{\tfrac{1}{2}}

\newcommand{\R}{\mathbb{R}}

\newcommand{\sign}{\operatorname{sign}}
\newcommand{\flatten}{\operatorname{flatten}}

\newcommand{\trace}{\operatorname{tr}}

\newcommand{\Id}{\mathbf{I}}

\newcommand{\abs}[1]{\vert {#1} \vert}
\newcommand{\norm}[1]{\Vert {#1} \Vert}

\def\vg{{\bm{g}}}

\def\vm{{\bm{m}}}

\def\vs{{\bm{s}}}
\def\vt{{\bm{t}}}
\def\vu{{\bm{u}}}
\def\vv{{\bm{v}}}
\def\vw{{\bm{w}}}
\def\vx{{\bm{x}}}
\def\vy{{\bm{y}}}

\def\mA{{\bm{A}}}

\def\mG{{\bm{G}}}

\def\mL{{\bm{L}}}
\def\mM{{\bm{M}}}

\def\mR{{\bm{R}}}

\def\mT{{\bm{T}}}
\def\mU{{\bm{U}}}
\def\mV{{\bm{V}}}
\def\mW{{\bm{W}}}
\def\mX{{\bm{X}}}

\def\mSigma{{\bm{\Sigma}}}

\newcommand{\ornament}{\vspace{1.5ex}\noindent\hfil{{\pgfornament[width=0.3\columnwidth]{82}}}\vspace{2ex}
}

\renewcommand*{\backrefalt}[4]{\ifcase #1 No citations. \or Cited on page #2. \else Cited on pages #2. \fi}

  \newtheorem{myproposition}{Proposition}

  \newtheorem{mydefinition}{Definition}

\SetAlCapSkip{1em}
\SetKwComment{Comment}{\small$\#$ }{}

\title[Old Optimizer, New Norm: An Anthology]{Old Optimizer, New Norm:\\An Anthology}

\optauthor{%
\Name{Jeremy Bernstein} \Email{jbernstein@mit.edu}\\
\Name{Laker Newhouse} \Email{lakern@mit.edu}\\
\addr MIT CSAIL, United States}

\begin{document}

\maketitle

\begin{abstract}
Deep learning optimizers are often motivated through a mix of convex and approximate second-order theory. We select three such methods---Adam, Shampoo and Prodigy---and argue that each method can instead be understood as a squarely first-order method without convexity assumptions. In fact, after switching off exponential moving averages, each method is equivalent to \textit{steepest descent} under a particular \textit{norm}. By generalizing this observation, we chart a new design space for training algorithms. Different operator norms should be assigned to different tensors based on the role that the tensor plays within the network. For example, while linear and embedding layers may have the same weight space of $\R^{m\times n}$, these layers play different roles and should be assigned different norms. We hope that this idea of carefully metrizing the neural architecture might lead to more stable, scalable and indeed faster training.
\end{abstract}

\section*{Prologue}

Deep learning optimizers are often motivated from the perspectives of convex and approximate second-order theory. These theoretical frameworks have been used to inspire algorithmic ideas, as well as providing means to analyse the convergence of various optimizers. However, we believe---and will attempt to demonstrate---that there is a wealth of untapped algorithmic opportunity in the simpler realm of exact first-order theory without convexity assumptions.

To make our case, we choose three optimizers that were originally analysed under convex or approximate second-order theory: Adam, Shampoo and Prodigy. After disabling their exponential moving averages (EMA), we show that each algorithm admits a parsimonious theoretical explanation as a variant of \textit{steepest descent} under a certain norm. EMA can then be thought of as ``smoothing out'' the algorithm, or making it more robust to mini-batch noise, although nailing down the precise role of EMA is perhaps still an open problem.

\begin{figure}
    \centering
    \includegraphics[width=\linewidth,trim={0 3.2cm 0 0},clip]{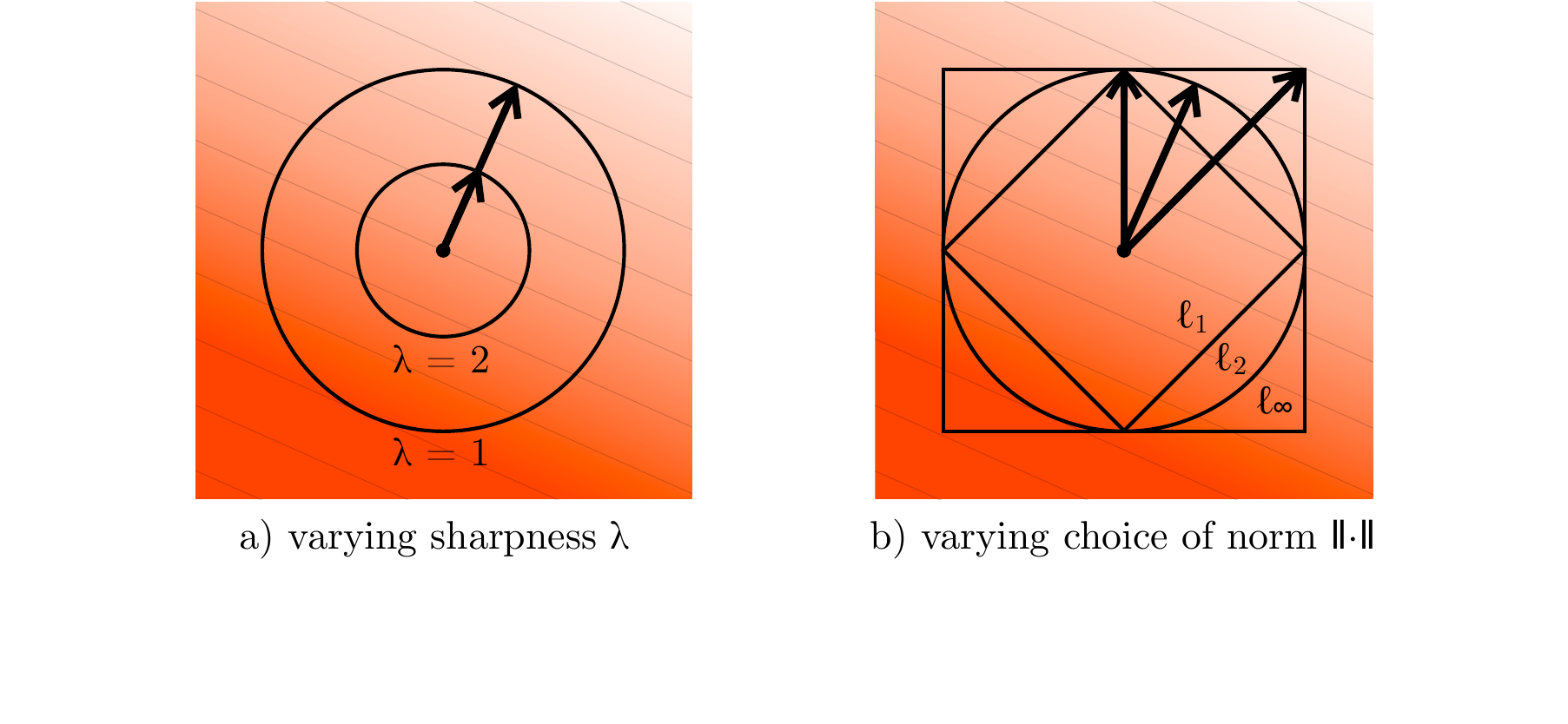}
    \vspace{-1em}
    \caption{Steepest descent considers the problem of minimizing a linear functional under a quadratic penalty: $\argmin_{\Delta \vw \in \R^n} \left[\vg^\top \Delta \vw + \frac{\lambda}{2} \, \norm{\Delta \vw}^2 \right]$ for $\vg \in \R^n$. Here we show how the solution varies with the sharpness $\lambda > 0$ and the choice of norm $\norm{\cdot}$. We overlay different norm balls on top of a linear color gradient, and use arrows to denote the solution, meaning the member of the norm ball that ``minimizes the color''. a) Increasing the sharpness decreases the size of the solution vector. b) Changing the norm can change the direction of the solution vector. For different $\ell_p$ norms, the solution direction changes because the gradient is not axis-aligned. In practice, we should pick the sharpness and norm to fit the geometry of our loss.}
    \label{fig:contours}
\end{figure}

By steepest descent, we mean the procedure of choosing a weight update $\Delta \vw$ to minimise a local quadratic model of the loss function $\el$ of the form $\el(\vw) + \nabla_\vw \el(\vw)^\top\Delta\vw + \frac{\lambda}{2}\cdot \norm{\Delta \vw}^2$, visualized in \cref{fig:contours}. Crucially, the \textit{sharpness parameter} $\lambda$ and \textit{norm} $\norm{\cdot}$ are chosen a priori, without touching an (approximate) Hessian during training. As such, we consider steepest descent to be a squarely first-order method and not an (approximate) second-order method.

Throughout the anthology, we rely on a dual description of steepest descent:
\begin{myproposition}[Steepest descent]\label{prop:steepest} For any $\vg \in \R^n$ thought of as ``the gradient'' and any $\lambda \geq 0$ thought of as ``the sharpness'', and for any norm $\norm{\cdot}:\R^n\to\R$ with dual norm $\norm{\cdot}^\dagger$:
\begin{align}\label{eq:dual-steepest}
    \argmin_{\Delta \vw \in \R^n} \left[\vg^\top \Delta \vw + \frac{\lambda}{2} \, \norm{\Delta \vw}^2 \right] = - \frac{\norm{\vg}^\dagger}{\lambda} \cdot \argmax_{\norm{\vt}=1} \vg^\top \vt.
\end{align}
\end{myproposition}
\cref{eq:dual-steepest} separates the solution of the steepest descent problem into two pieces: first computing the \textit{step size} as the dual norm of the gradient divided by the sharpness, and second solving for the \textit{step direction} as the unit vector that maximizes the inner product with the gradient. The proof of this proposition is given in \cref{proof:steepest}.

Of course, the art of steepest descent lies in choosing a norm $\norm{\cdot}$ and a sharpness $\lambda$ suited to the optimization problem at hand. While it may be possible to turn this art into a science \citep{modula}, that ambition is beyond the scope of this anthology. Here we point out that past methods do implicitly make decisions about norms, and in a somewhat haphazard manner. In fact, they implicitly assign different \textit{induced matrix norms} to the network layers:

\begin{mydefinition}[Induced 
operator norm]\label{def:induced} Given a matrix $\mM\in\R^{d_\out \times d_\inn}$ and two normed vector spaces $(\R^{d_\inn},\norm{\cdot}_\alpha)$ and $(\R^{d_\out},\norm{\cdot}_\beta)$, the ``$\alpha$ to $\beta$'' induced operator norm is given by:
\begin{equation}
    \norm{\mM}_{\alpha\to\beta} = \max_{\substack{\vx \in \R^{d_\inn}}} 
    \frac{\norm{\mM\vx}_\beta}{\norm{\vx}_\alpha}.
\end{equation}
\end{mydefinition}
\cref{def:induced} tells us that by varying the choice of vector norms $\norm{\cdot}_\alpha$ and $\norm{\cdot}_\beta$, we can induce a large family of matrix norms. In turn, this implies a correspondingly large family of steepest descent optimizers. By foregrounding this issue, we hope that algorithm designers may develop more suitable optimizers by becoming more intentional about their choice of norm.

\section{Adam as Steepest Descent under the Max-of-Max Norm}
\label{sec:adam}

\lettrine{A}{dam} is a widely used deep learning optimizer: the original paper of \citet{kingma_adam:_2015} now has well over 100,000 citations. Adam has been motivated in various ways, including through convex analysis \citep{kingma_adam:_2015} and as an approximate second-order method \citep{sun-and-spall}. However, there have been efforts to build a more direct understanding of Adam: for instance, with exponential moving averages (EMA) switched off, Adam is just \textit{sign gradient descent} \citep{pmlr-v80-balles18a,signum}, which is equivalent to steepest descent under the infinity norm \citep{spectral-descent-2}. In this story, we connect Adam to a certain ``max-of-max'' norm, showing how Adam respects the tensor structure of a neural network in a very particular way.

To begin, we review how Adam connects to sign gradient descent. Ignoring bias corrections and numerical stabilizations, Adam is given by the following system of updates:
\begin{align}
    \vm_t &= \beta_1 \cdot \vm_{t-1} + (1 - \beta_1) \cdot \vg_t, \\
    \vv_t &= \beta_2 \cdot \vv_{t-1} + (1 - \beta_2) \cdot \vg_t^2, \\
    \vw_{t+1} &= \vw_t - \eta \cdot \vm_t / \sqrt{\vv_t},
\end{align}
where $t$ denotes the time step, $\vg_t \in \R^n$ the gradient vector and $\eta > 0$ the step size. The EMA time scales of the first gradient moment $\vm_t$ and second moment $\vv_t$ are set by $0 \leq \beta_1, \beta_2 < 1$. All operations are conducted entry-wise. If we switch off EMA by setting $\beta_1 = \beta_2 = 0$, the Adam updates reduce to just sign gradient descent:
\begin{align}
    \vw_{t+1} &= \vw_t - \eta \cdot \vg_t / \sqrt{\vg_t^{2}} \label{eq:funky-adam} \\
    &= \vw_t - \eta \cdot \sign(\vg_t) \label{eq:sign-adam}.
\end{align}

This connection to sign descent should not be surprising since Adam, published in 2015, builds on the RMSprop optimizer that \citet{tieleman_rmsprop_2012} already called ``the mini-batch version of just using the sign of the gradient''. And RMSprop itself built on the RPROP optimizer \citep{rprop}, which also uses gradient signs.

Still, \textit{why should using the sign of the gradient be a good idea in deep learning?} In search of a motivation, we might consider that sign descent solves the problem of steepest descent under the vector $\ell_\infty$ norm, $\norm{\vv}_\infty \defeq \max_i \abs{\vv_i}$ \citep{spectral-descent-2, spectral-descent-1, xie2024implicit}:
\begin{myproposition}[Sign descent as steepest descent under the infinity norm]\label{prop:sign-descent}
    \\For any gradient vector $\vg\in\R^n$ and sharpness $\lambda > 0$, it holds that:
    \begin{equation}
        \argmin_{\Delta \vw \in \R^n} \left[ \vg^\top \Delta \vw + \frac{\lambda}{2} \, \norm{\Delta \vw}_\infty^2 \right] = -\frac{\norm{\vg}_1}{\lambda} \, \sign(\vg).
    \end{equation}
\end{myproposition}
In words, the vector that minimizes a linear functional under an infinity norm penalty is a scalar multiple of a sign vector. The proof is given in \cref{proof:sign-descent}. 

While this connection between Adam, sign descent and steepest descent is perhaps cute, it does not answer a basic question: \textit{Why does the vector $\ell_\infty$ norm have anything to do with neural network training?} In particular, taking the weight space to be $\R^n$ equipped with the simple infinity norm seems to ``throw away'' the fact that the weight space of a neural network is built in a structured way out of layers of matrices (and perhaps other tensors).

To resolve this conundrum, we suggest that in fact the vector $\ell_\infty$ norm on the flattened weight space doesn't have anything to do with deep learning. Instead, there is a coincidence at play. The $\ell_\infty$ norm enjoys a special property summarized by the slogan ``a max of a max is a max''. To see this, consider a neural network with a list of $L$ weight matrices $\mW_1, \dots, \mW_L$. Let $\mathrm{row}_r(\mW_l)$ denote the $r$th row of the $l$th weight matrix, and let $\vw = \flatten(\mW_1, \dots, \mW_L) \in \R^n$ denote the full flattened weight vector. Then we have that:
\begin{equation}\label{eq:coincidence}
    \norm{\vw}_\infty = \max_{l} \max_r \norm{ \mathrm{row}_r(\mW_l) }_\infty = \max_{l} \norm{\mW_l}_{\ell_1\to\ell_\infty},
\end{equation}
where the second equality follows via \cref{prop:tractable-norms}. In words, the infinity norm of the flattened weight vector coincides with the largest $\ell_1$ to $\ell_\infty$ operator norm of the layers. So \cref{eq:coincidence} connects the \textit{unstructured} space of the flattened weight vector to the \textit{structured} space of the list of weight matrices. We refer to the object $\max_{l} \norm{\mW_l}_{\ell_1\to\ell_\infty}$ as the ``max-of-max norm''. And sign descent emerges as steepest descent under this norm:
\begin{myproposition}[Sign descent as steepest descent under the max-of-max norm]\label{prop:structural-sign-descent}
    \\For any list of gradient matrices $\mG_1,...,\mG_L$ and any sharpness $\lambda > 0$, consider the problem:
    \begin{equation}\label{eq:max-to-max}
        \argmin_{\Delta \mW_1,...,\Delta \mW_L} \left[ \sum_{l=1}^L \langle\mG_l, \Delta \mW_l\rangle + \frac{\lambda}{2} \max_{l=1}^L \norm{\Delta\mW_l}_{\ell_1\to\ell_\infty}^2 \right],
    \end{equation}
    where $\langle\cdot, \cdot\rangle$ denotes the Frobenius inner product, and $\Delta \mW_l$ has the same shape as $\mG_l$. For step size $\eta = \frac{1}{\lambda}\sum_{l=1}^L \norm{\mG_l}_{\ell_1\to\ell_\infty}^\dagger$, where $\dagger$ denotes the dual norm, \cref{eq:max-to-max} is solved by:
    \begin{equation}\label{eq:sign-descent-layerwise}
        \Delta \mW_l = - \eta \cdot \sign(\mG_l) \qquad \text{ for each layer } l=1,...,L.
    \end{equation}
\end{myproposition}
In words, the matrix-aware steepest descent problem of \cref{eq:max-to-max} is solved by layerwise sign descent as given in \cref{eq:sign-descent-layerwise}. This observation---that sign descent updates are implicitly doing \textit{per-matrix gradient normalization}---may be a major reason that Adam, sign descent and Lion \citep{chen2023symbolic} outperform vanilla gradient descent in large language model training \citep{Zhao2024Deconstructing, modula}. The proof is given in \cref{proof:structural-sign-descent}.

\ornament

All told, this story has shown that Adam without EMA is sign descent and that, coincidentally, sign descent solves two different steepest descent problems: one on the flattened weight space, and one that is aware of the matrix structure of neural architecture. But, at the end of this story, questions linger. \textit{Why does the $\ell_1$ to $\ell_\infty$ induced operator norm rear its head? What does it have to do with deep learning? Aren't there other induced operator norms on matrices we could equally well consider?} For answers to these questions, dear reader, you'll have to wait for our next story... a story about Shampoo!

\clearpage
\section{Shampoo as Steepest Descent under the Spectral Norm}
\label{sec:shampoo}

\lettrine{N}{ow}, dear reader, we turn our attention to Shampoo \citep{unified-shampoo,Gupta2018ShampooPS}. A variant of the Shampoo optimizer won the external tuning track of the \textit{2024 AlgoPerf: Training Algorithms} competition \citep{Dahl2023AlgoPerf}. While the method was originally motivated as a generalization of the AdaGrad convex optimizer \citep{Duchi2011AdaptiveSM} to tensor spaces, more recent work casts Shampoo as an approximate second-order method \citep{Anil2020ScalableSecondOrder,Morwani2024NewPerspective}. We will show that Shampoo---with accumulation disabled---is steepest descent under the max spectral norm over layers.

To begin, we show that Shampoo updates, without accumulation, are semi-orthogonal matrices. At time step $t$ and for each layer, Shampoo collects the gradient matrix $\mG_t$ and makes the following update to the weight matrix $\mW_t$:
\begin{align}
    \mL_t &= \mL_{t-1} + \mG_t \mG_t^T, \label{eq:L}\\
    \mR_t &= \smash{\mR_{t-1} + \mG_t^T \mG_t}, \label{eq:R}\\
    \mW_{t+1} &= \smash{\mW_t - \eta \cdot \mL_t^{-\sfrac{1}{4}} \mG_t \mR_t^{-\sfrac{1}{4}}}.
\end{align}
All operations, including the inverse fourth roots, are matrix operations. The accumulators $\mL_t$ and $\mR_t$ are referred to as the ``left and right pre-conditioners''. Practitioners usually replace the simple sums in \cref{eq:L,eq:R} with EMAs \citep{Shi2023DistributedShampoo}. If we disable the accumulation, setting $\mL_t = \mG_t\mG_t^\top$ and $\mR_t = \mG_t^\top \mG_t$, Shampoo reduces to:
\begin{align}
    \mW_{t+1} &= \smash{\mW_t - \eta \cdot (\mG_t\mG_t^\top)^{-\sfrac{1}{4}} \,\mG_t\, (\mG_t^\top\mG_t)^{-\sfrac{1}{4}}} \label{eq:funky-shampoo}\\
    &= \smash{\mW_t - \eta \cdot \mU_t \mV_t^\top}, \label{eq:orthog-shampoo}
\end{align}
where \cref{eq:orthog-shampoo} is reached by substituting the reduced singular value decomposition (SVD) of the gradient $\mG_t = \mU_t \mSigma_t \mV_t^\top$ into \cref{eq:funky-shampoo}. Notice that there is a direct parallel between \cref{eq:funky-adam,eq:sign-adam} for Adam and \cref{eq:funky-shampoo,eq:orthog-shampoo} for Shampoo. So, Shampoo without accumulation makes a \textit{semi-orthogonal weight update}. In fact:
\begin{myproposition}[Projection to the closest semi-orthogonal matrix]\label{prop:projection} Consider the\\semi-orthogonal matrices $\mathcal{O}_{m \times n} \defeq \left\{ \mA \in \R^{m \times n} : \mA\mA^\top = \Id_{m} \text{ or } \mA^\top \mA = \Id_{n}\right\}$ and let $\norm{\cdot}_F$ denote the Frobenius norm. For any matrix $\mG \in \R^{m \times n}$ with reduced SVD $\mG = \mU \mSigma \mV^\top$:
\begin{equation}
    \argmin_{\mA \in \mathcal{O}_{m \times n}} \norm{\mA - \mG}_F = \mU \mV^\top,
\end{equation}
where the minimizer $\mU\mV^\top$ is unique if and only if the matrix $\mG$ has full rank.
\end{myproposition}
So, Shampoo without accumulation projects the gradient matrix to the \textit{closest semi-orthogonal matrix in Frobenius norm}. The proof is in \cref{proof:projection}. \textit{Why might this be a good idea}, you ask? Well, for one thing, it's steepest descent---this time under the maximum \textit{spectral norm} $\norm{\cdot}_{\ell_2 \to \ell_2
}$ (\cref{def:induced}) over all the matrices in the network:

\begin{myproposition}[Shampoo as steepest descent under the spectral norm]\label{prop:shampoo-steepest} 
    For any list of gradient matrices $\mG_1,...,\mG_L$ and any sharpness $\lambda > 0$, consider the problem:
    \begin{equation}\label{eq:max-of-spectral}
        \argmin_{\Delta \mW_1,...,\Delta \mW_L} \left[ \sum_{l=1}^L \langle\mG_l, \Delta \mW_l\rangle + \frac{\lambda}{2} \, \max_{l=1}^L \norm{\Delta\mW_l}_{\ell_2\to\ell_2}^2 \right],
    \end{equation}
    where $\langle\cdot, \cdot\rangle$ denotes the Frobenius inner product and $\Delta \mW_l$ has the same shape as $\mG_l$. Suppose that $\mG_l$ has reduced SVD given by $\mG_l = \mU_l \mSigma_l \mV_l^\top$ for each $l=1,...,L$. Then \cref{eq:max-of-spectral} is solved with a step size $\eta = \frac{1}{\lambda}\sum_{l=1}^L \trace \mSigma_l$ and an update:
    \begin{equation}\label{eq:shampoo-layerwise}
        \Delta \mW_l = -\eta \cdot \mU_l \mV_l^\top \quad \text{ for each } l=1,...,L.
    \end{equation}
    This solution for $\Delta \mW_l$ is unique if and only if the matrix $\mG_l$ is of full rank.
\end{myproposition}
The proof is given in \cref{proof:shampoo-steepest}. A novelty of this proposition in contrast to prior work on stochastic spectral descent \citep{spectral-descent-2, spectral-descent-1} is our use of a max norm over layers to handle the multi-layer case. However, our main contribution here is to draw the connection between \cref{prop:shampoo-steepest} and Shampoo as in \cref{eq:funky-shampoo,eq:orthog-shampoo}.

So, Shampoo without accumulation is steepest descent under the spectral norm. \textit{Why might this be a good idea in deep learning?} The idea that we wish to advance is that one can derive upper bounds on the loss of machine learning models in terms of spectral norms. Here we present the simplest possible example: a linear model and the square loss.

\begin{myproposition}[Bounding the square loss of a linear predictor]\label{prop:majorization} Consider a matrix $\mW \in \R^{d_\mathrm{out}\times d _\mathrm{in}}$ that we shall think of as a linear predictor mapping an input $\vx \in \R^{d_\mathrm{in}}$ to an output $\vy = \mW \vx \in \R^{d_\mathrm{out}}$. Given a dataset of $n$ samples $\mathcal{D} = \{(\vx_1,\vy_1), ..., (\vx_n,\vy_n)\}$, where the $i$th input is normalized such that $\|\vx_i\|_2 = \sqrt{d_\inn}$, we can construct the ``square loss'':
\begin{equation}
    \el(\mW) \defeq \frac{1}{2n}\sum_{i=1}^n \frac{1}{d_\out}\norm{\vy_i - \mW \vx_i}_2^2.
\end{equation}
Then, for any matrix $\Delta\mW \in \R^{d_\mathrm{out}\times d _\mathrm{in}}$ thought of as a weight update, it holds that:
\begin{equation}\label{eq:square-majorization}
    \el(\mW + \Delta \mW) \leq \el(\mW) + \langle\nabla_\mW \el(\mW), \Delta \mW\rangle + \half \cdot \tfrac{d_\inn}{d_\out}\cdot\norm{\Delta \mW}_\mathrm{\ell_2\to\ell_2}^2,
\end{equation}
where $\langle\cdot,\cdot\rangle$ is the Frobenius inner product.
\end{myproposition}
In words: the square loss of a linear predictor admits an upper bound that is quadratic in the spectral norm of the weight perturbation. Choosing the weight perturbation to minimize this upper bound is precisely steepest descent under the spectral norm! The proof is given in \cref{proof:majorization}. This optimizer design pattern, which starts by deriving an upper bound on the loss (as in \cref{prop:majorization}) and then minimizes it (as in \cref{prop:shampoo-steepest}), is known generally as majorization-minimization \citep{mm}. It is an exact and first-principles design pattern, without Hessian approximations or appeals to convex theory. This design pattern is used extensively by \citet{spectral-descent-2, spectral-descent-1} to design optimizers for restricted Boltzmann machines and discrete graphical models. Generalizing the pattern to arbitrary network architectures and loss functions requires more advanced machinery \citep{agd-2023,streeter2023universal,modula}.

\ornament

And so, dear reader, we have reached the end of our second story. We have shown that Shampoo without accumulation corresponds to projecting the gradient matrix to the closest semi-orthogonal matrix, which solves the problem of steepest descent under the spectral norm. And we showed how steepest descent under the spectral norm emerges from upper bounding the square loss of a linear predictor. This perspective, of viewing Shampoo as a (smoothed out) projection to the space of semi-orthogonal matrices, grounds the algorithm in a prior literature on spectral descent \citep{spectral-descent-2,spectral-descent-1,spectral-descent-3}. And in \cref{app:shampoo}, we discuss how it might unlock new means for computing the Shampoo updates.

We summarize our first two stories in \cref{tab:weight_update_rules}. And we still have one more left to tell...

\renewcommand{\arraystretch}{1.5}
\begin{table}
    \centering
    \begin{tabular}{cclcc}
        \toprule
        \textbf{Domain} & \textbf{Norm} & \hspace{1.3em}\textbf{Solution} & \textbf{Optimizer} & \textbf{Cousin}\\
        \midrule
        $\R^n$ & Euclidean $\ell_2$ & $\displaystyle\Delta \vw = -\tfrac{\norm{\vg}_2}{\lambda} \, \tfrac{\vg}{\norm{\vg}_2}$ & vanilla gradient descent & SGD \\
        $\R^n$ & infinity $\ell_\infty$ & $\displaystyle\Delta \vw = -\tfrac{\norm{\vg}_1}{\lambda} \sign(\vg)$ & sign descent & Adam\\
        \midrule
        $\R^{m\times n}$ & Frobenius $S_2$ & $\displaystyle\Delta \mW = -\tfrac{\norm{\mG}_F}{\lambda} \, \tfrac{\mG}{\norm{\mG}_F}$ & vanilla gradient descent & SGD\\
        $\R^{m\times n}$ & spectral $S_\infty$ & $\displaystyle\Delta \mW = -\tfrac{\trace \mSigma}{\lambda}\,\mU\mV^\top$ & spectral descent & Shampoo\\
        \bottomrule
    \end{tabular}
    \caption{Popular optimizers are related to steepest descent under different norms. For vector-valued optimization problems, we consider the steepest descent problem $\smash{\argmin_{\Delta \vw} \vg^\top \Delta \vw + \frac{\lambda}{2}\cdot \norm{\Delta \vw}^2}$. For matrix-valued problems, we consider $\smash{\argmin_{\Delta \mW} \, \langle\mG, \Delta \mW\rangle + \frac{\lambda}{2}\cdot \norm{\Delta \mW}^2}$, where $\langle\cdot,\cdot\rangle$ is the Frobenius inner product. We list the solution for different vector $\ell_p$ norms and Schatten $S_p$ norms. The Schatten $S_p$ norm of a matrix returns the $\ell_p$ norm of its vector of singular values. Finally, $\mG = \mU \mSigma \mV^\top $ is the reduced singular value decomposition of the gradient.}
    \label{tab:weight_update_rules}
\end{table}
\renewcommand{\arraystretch}{1}

\clearpage
\section{Prodigy: Automatically Computing the Escape Velocity}
\label{sec:prodigy}

\lettrine{F}{or} our final story, we speak of Prodigy \citep{Mishchenko2023Prodigy}. The Prodigy optimizer falls amid a series of recent works \citep{pmlr-v202-defazio23a,khaled2023dowg,Ivgi2023DoGIS} that attempt to apply convex theory to design and analyse deep learning optimizers that do not require tuning. In contrast, we argue that Prodigy (without EMA) is but another example of steepest descent, where instead of using the step size $\eta=\norm{\vg}^\dagger/\lambda$ from \cref{prop:steepest}, Prodigy uses a heuristic to automatically warm up to a good step size. This demonstrates the value of \cref{prop:steepest} for disentangling the optimizer design problem. If one knows a good norm $\norm{\cdot}$ but is ignorant of the sharpness parameter $\lambda$, then one may obtain the step direction by solving $\argmax_{\norm{\vt}=1} \vg^\top \vt$ from \cref{prop:steepest}, while using another means to find a good step size.

Then let us make our case. We focus on Algorithm 3 in the Prodigy paper, since this is the version used in their experiments. We first show that with EMA switched off, Prodigy implements sign gradient descent with a step size that warms up automatically. Ignoring the numerical stabilization and learning rate schedule, Prodigy is given by:
\begin{align}
    \vm_t &= \beta_1 \cdot \vm_{t-1} + (1 - \beta_1) \cdot \eta_t \,\vg_t, \\
    \vv_t &= \beta_2 \cdot \vv_{t-1} + (1 - \beta_2) \cdot \eta_t^2 \,\vg_t^2, \\
    r_t &= \sqrt{\beta_2} \cdot r_{t-1} + (1-\sqrt{\beta_2}) \cdot \eta_t^2 \, \vg_t^\top(\vw_0 - \vw_t), \label{eq:r-prodigy} \\
    \vs_t &= \sqrt{\beta_2} \cdot \vs_{t-1} + (1-\sqrt{\beta_2}) \cdot \eta_t^2 \, \vg_t, \label{eq:s-prodigy} \\
    \eta_{t+1} &= \max\left(\eta_t, \tfrac{r_t}{\norm{\vs_t}_1}\right), \label{eq:d-prodigy} \\
    \vw_{t+1} &= \vw_t - \eta_t \cdot \vm_t / \sqrt{\vv_t},
\end{align}
where $t$ denotes the time step and $\vg_t \in \R^n$ the gradient vector. While this system of updates may seem intimidating, if we switch off EMA by setting $\beta_1 = \beta_2 = 0$, the Prodigy updates simplify dramatically to just sign gradient descent with a dynamical step size as follows:
\begin{align}
    \eta_{t+1} &= \max\left(\eta_t, \tfrac{\vg_t^\top(\vw_0 - \vw_t)}{\norm{\vg_t}_1}\right), \label{eq:adaptive-prodigy}
    \\
    \vw_{t+1} &= \vw_t - \eta_t \cdot \sign (\vg_t).
    \label{eq:sign-prodigy}
\end{align}
But \cref{prop:sign-descent} showed that sign descent is steepest descent under the infinity norm. Therefore \cref{eq:adaptive-prodigy,eq:sign-prodigy} prove our claim that Prodigy without EMA is steepest descent, although with a dynamically chosen step size denoted $\eta_t$.

All that remains is to understand the dynamical rule, given by \cref{eq:adaptive-prodigy}, for choosing the step size $\eta_t$. We shall argue that this dynamical rule can be understood to approximate a heuristic algorithm for achieving, but not exceeding, what we shall call \textit{escape velocity}:
\begin{itemize}
    \item Choose a very small initial step size $\eta_0$---small enough to be \textit{a priori} sure that $\eta_0 \ll \eta_\star$, where $\eta_\star$ denotes \textit{escape velocity}: the unknown but optimal initial step size;
    \item At each step, check if the weights $\vw_t$ have escaped the linearization of the loss around the initial weights $\vw_0$---if not, double the step size according to $\eta_{t+1} = 2 \times \eta_t$;
    \item Once the weights $\vw_t$ have escaped the initial linearization, stop increasing the step size. We say that the step size $\eta_t$ has reached escape velocity $\eta_\star$.
\end{itemize}
The rationale behind this procedure is that if we knew the optimal initial step size $\eta_\star$, then the weights should escape the initial linearization of the loss in a single step. Formally, the directional derivative $(\vw_1 - \vw_0)^\top \vg_1$ must vanish if the step size is chosen optimally \citep{cauchy1847methode}. If the directional derivative in the direction of the first weight update is still negative $(\vw_1 - \vw_0)^\top \vg_1 < 0$, then we could have taken a larger step. Said another way, we can use the \textit{angle} that the gradient $\vg_1$ makes with the change in weights $\vw_1 - \vw_0$ to tell us whether or not we should increase the step size. Notice that procedure has no reliance on convexity. 

With this in mind, let us massage Prodigy's step size update (\cref{eq:adaptive-prodigy}) as follows:
\begin{align}\label{eq:rule}
    \eta_{t+1} = \max\left(\eta_t, \tfrac{\vg_t^\top(\vw_0 - \vw_t)}{\norm{\vg_t}_1}\right) = \max\left(\eta_t, \tfrac{\norm{\vg_t}_2}{\norm{\vg_t}_1}\times \norm{\vw_t - \vw_0}_2 \times \cos\theta\right),
\end{align}
where $\theta$ denotes the angle between the gradient $\vg_t$ and the difference in weights $\vw_0 - \vw_t$. To help make sense of this expression, we make two assumptions:
\begin{enumerate}
    \item The gradient is a ``dense'' vector in $\R^n$, meaning that $\norm{\vg_t}_2 / \norm{\vg_t}_1 \approx  1/ \sqrt{n}$; \label{ass:1}
    \item $\vw_t$ is still close enough to the initialization $\vw_0$ that $\cos\theta \approx 1$. \label{ass:2}
\end{enumerate}
Under these assumptions, \cref{eq:rule} becomes just $\eta_{t+1} \approx \max\left(\eta_t, \norm{\vw_t - \vw_0}_{\rms}\right)$, where the root mean square (RMS) norm is defined via $\norm{\cdot}_\rms \defeq \tfrac{1}{\sqrt{n}}\,\norm{\cdot}_2$. Combined with \cref{eq:sign-prodigy}, this allows us to estimate the size of the weight change at step $t+1$:
\begin{equation*}
    \norm{\vw_{t+2} - \vw_{t+1}}_\rms = \eta_{t+1} \cdot \norm{\sign(\vg_t)}_\rms \approx \max\left(\eta_t, \norm{\vw_t - \vw_0}_{\rms}\right) \geq \norm{\vw_t - \vw_0}_{\rms},
\end{equation*}
where we have used the fact that a sign vector has unit RMS norm. In words, while assumptions (\ref{ass:1}) and (\ref{ass:2}) hold, the step size at time $t+1$ is equivalent to the whole progress up to step $t$. This suggests exponential growth in the step size that continues until assumption (\ref{ass:2}) breaks, which we think of as the step size reaching the escape velocity $\eta_\star$

Now we wish to point out that this procedure is just one amongst a space of line search methods that one might consider \citep{armijo1966,rprop,kenneweg2024}. For instance, Prodigy's decision to only let $\eta_t$ increase and never decrease could be sub-optimal. And the decision to measure the angle between the gradient and the weight difference $\vw_t - \vw_0$ has alternatives. One could instead use the most recent weight difference $\vw_t - \vw_{t-1}$. Lastly, in place of relying on the norm ratio $\norm{\vg}_2/\norm{\vg_1}$ to implicitly convert the $\ell_2$ norm $\norm{\vw_t -\vw_0}_2$ into the RMS norm $\norm{\vw_t -\vw_0}_\rms$, one could consider a more explicit method. For instance, we found a rule akin to $\eta_{t+1} = \eta_t \times (1 + \cos\theta)$ to work well in some preliminary experiments.

\ornament

Our time grows short, dear reader, and our third story draws to an end. We have argued that Prodigy without EMA is sign descent---an example of steepest descent---with a particular mechanism for warming up the step size. Starting with a tiny initial step size, Prodigy multiplicatively increases the step size until the weights escape the initial locally linear region of the loss. Prodigy's step size adjustment is based on the angle between the gradient and the total weight change. This is a form of online line search. This highlights that once one has chosen a norm, the steepest descent framework allows freedom to estimate the step size in various different ways.

\clearpage
\section*{Epilogue}
\label{sec:epilogue}

This anthology has presented new ways of understanding old optimizers. \cref{prop:steepest} decouples the optimizer design problem into two pieces: first choosing a norm and second finding a step size. This design space is already broad. We have argued that Adam chooses the infinity norm (\cref{prop:sign-descent}) or equivalently the max-of-max norm (\cref{prop:structural-sign-descent}), which respects a layered matrix structure. Shampoo chooses the spectral norm (\cref{prop:shampoo-steepest}). Prodigy chooses the same norm as Adam, and then uses a heuristic to automatically warm up to a good step size, as in \cref{eq:adaptive-prodigy}, which we term \textit{reaching escape velocity}.

Through the lens of steepest descent, the decisions that Adam, Shampoo and Prodigy make may seem arbitrary. In fact, we think that they \textit{are} somewhat arbitrary. And there may be more principled ways to make these decisions. To demonstrate this point, we now introduce a tool called the modular norm \citep{modula} and its corresponding steepest descent algorithm. The modular norm generalizes the norms that appeared in \cref{prop:structural-sign-descent} for Adam and \cref{prop:shampoo-steepest} for Shampoo. Formally:
\begin{myproposition}[Steepest descent under the modular norm]\label{prop:steepest-modular} Given scalar coefficients $s_1, \dots, s_L > 0$ and norms $\norm{\cdot}_1, \dots, \norm{\cdot}_L$, we define the modular norm as the mapping:
\begin{equation}
     \mW_1, \dots, \mW_L \mapsto \max\left\{s_1 \norm{\mW_1}_1, \dots, s_L \norm{\mW_L}_L \right\}.
\end{equation}
The corresponding steepest descent problem is given by:
\begin{equation}\label{eq:steepest-modular}
    \argmin_{\Delta \mW_1, \dots, \Delta \mW_L} \left[ \sum_{l=1}^L \langle \mG_l, \Delta \mW_l \rangle + \frac{\lambda}{2} \max_{l=1}^L s_l^2 \norm{\Delta\mW_l}_l^2 \right],
\end{equation}
where $\langle\cdot, \cdot\rangle$ denotes the Frobenius inner product, and for each $l=1,...,L$ the two matrices $\Delta \mW_l$ and $\mG_l$ are of the same shape. If we define the global step size $\eta = \frac{1}{\lambda}\sum_{k=1}^L \frac{1}{s_k}\norm{\mG_k}_k^\dagger$, then the solution to \cref{eq:steepest-modular} is given by:
\begin{align}
    \Delta \mW_l = - \frac{\eta}{s_l}\cdot \argmax_{\norm{\mT_l}_l=1} \, \langle\mG_l, \mT_l\rangle \quad\text{ for each layer } l = 1,...,L.
\end{align}
\end{myproposition}
In words, steepest descent under the modular norm updates each layer in a direction informed by that layer's norm and with a global step size computed as a weighted sum of the dual norms of the gradients over layers. The proof of this proposition is given in \cref{proof:steepest-modular}.

When confronted with the modular norm, it's natural to ask how one should assign norms to layers. And there are so many norms to choose from! Beyond the familiar $\ell_2 \to \ell_2$ spectral norm, many other induced operator norms are computationally tractable:
\begin{myproposition}[$\ell_1\to\ell_p$ and $\ell_p \to \ell_\infty$ induced operator norms are tractable]\label{prop:tractable-norms} For a matrix $\mM\in\R^{m\times n}$ with $m$ rows $\{\mathrm{row}_i(\mM)\}_{i=1}^m$ and $n$ columns $\{\mathrm{col}_j(\mM)\}_{j=1}^n$, and $1\leq p \leq \infty$:
    \begin{align}
        \norm{\mM}_{\ell_1\to \ell_p} = \max_j \norm{\mathrm{col}_j(\mM)}_p; \qquad
        \norm{\mM}_{\ell_p\to \ell_\infty}= \max_i \norm{\mathrm{row}_i(\mM)}_{\frac{p}{p-1}}.
\end{align}
\end{myproposition}
In words, the $\ell_1\to\ell_p$ operator norm is the largest $\ell_p$ norm of the columns; the $\ell_p\to\ell_\infty$ operator norm is the largest dual $\ell_p$ norm over the rows. The proof is given in \cref{proof:tractable-norms}.

To assign a norm to a layer, we believe that one should consider the role that layer plays in the neural network. For instance, since linear layers are typically used to map to and from vectors with roughly unit RMS norm, it is appropriate to equip linear layers with the induced RMS to RMS operator norm \citep{my-spectral}, which resolves to a rescaled spectral norm. And since embedding layers map from one-hot vectors to vectors with roughly unit RMS norm, it is appropriate to equip embedding layers with the $\ell_1$ to RMS operator norm, which resolves to a rescaled $\ell_1$ to $\ell_2$ operator norm. So embedding layers and linear layers should be equipped with different norms despite the weight space being a matrix space in both cases. In short, the algorithm designer has freedom to choose input and output norms for layers that capture differences in how the layers are used; inducing the corresponding operator norm on the layer's weights provides control over how the optimizer learns representations.

We believe that picking the right norms could improve the speed and scalability of neural network training. We are seeing evidence that equipping neural network layers with better norms can lead to learning rate transfer across scale \citep{my-spectral,modula}. And since Shampoo won the external tuning track of the 2024 AlgoPerf competition \citep{Dahl2023AlgoPerf}, it is garnering interest as a fast training method. The second story in our anthology shows that Shampoo is closely connected to the spectral norm.

In conclusion, this work highlights a perspective on optimizer design as choosing two things: a \textit{norm} and a \textit{step size}. We have shown that three popular methods---Adam, Shampoo and Prodigy---fit within this perspective. We hope that researchers can design improved training algorithms by choosing norms and step sizes more intentionally.

\vspace{1em}

\hfill \textit{``Though this be madness, yet there is method in't.''}\\
\vspace{-2ex}
\hfill Hamlet

\section*{Acknowledgements}
We are grateful to Tim Large and Phillip Isola for invaluable discussions on the stories in this anthology. We also thank Jack Gallagher, Keller Jordan, Tongzhou Wang and Victor Butoi for very helpful conversations.

\clearpage
\bibliography{refs}

\newpage
\clearpage
\appendix

\section{Computational Strategies for Shampoo}
\label{app:shampoo}

Let $\mG\in\R^{m\times n}$ be a gradient matrix with reduced SVD $\mG = \mU \mSigma \mV^\top$. By \cref{eq:funky-shampoo,eq:orthog-shampoo}, the corresponding Shampoo update (with EMA disabled) is given by:
\begin{equation}\label{eq:shampoo-abridged}
    \Delta \mW = - \eta \cdot (\mG\mG^\top)^{-\sfrac{1}{4}} \,\mG\, (\mG^\top\mG)^{-\sfrac{1}{4}} = - \eta \cdot \mU \mV^\top.
\end{equation}
Here we list every means we know of computing or approximating this equation. First, we mention that $(\mG\mG^\top)^{-\sfrac{1}{4}} \,\mG\, (\mG^\top\mG)^{-\sfrac{1}{4}} = (\mG\mG^\top)^{-\sfrac{1}{2}} \,\mG = \mG\, (\mG^\top\mG)^{-\sfrac{1}{2}}$, so if one is willing to compute inverse matrix roots, one need only compute either $(\mG\mG^\top)^{-\sfrac{1}{2}}$ or $(\mG^\top\mG)^{-\sfrac{1}{2}}$, whichever has smaller dimension. With that said, to compute \cref{eq:shampoo-abridged}, one may:
\begin{enumerate}
    \item \textbf{Do the SVD.} Apply an SVD routine to compute $\mU$, $\mSigma$ and $\mV^\top$ and just discard $\mSigma$.
    \item \textbf{Do sketching.} Sketching is a randomized method \citep{Martinsson_Tropp_2020} that can be used to approximate the SVD. See, for instance, Sketchy \citep{sketchy} and spectral descent for deep learning \citep{spectral-descent}.
    \item \textbf{Do Newton iteration for inverse $p$th roots.} Inverse matrix roots such as $(\mG\mG^\top)^{-\sfrac{1}{2}}$ can be computed via Newton iteration \citep{lakic}. This is discussed in Chapter 7 of \citet{higham}'s book. And see \citet{Anil2020ScalableSecondOrder}'s paper.
    \item \textbf{Do Newton-Schulz iteration.} We developed a ``Newton-Schulz iteration'' for computing $\mU\mV^\top$, adapted from Equation 5.22 in \citet{higham}'s book. In short, if we set $\mX_0 = \mG / \norm{\mG}_{\ell_2 \to \ell_2}$ (or alternatively $\mX_0 = \mG / \norm{\mG}_F$) and iterate:
    \begin{equation}\label{eq:newton-schulz}
        \mX_{t+1} = \frac{3}{2} \cdot \mX_t - \frac{1}{2} \cdot \mX_t \mX_t^\top \mX_t,
    \end{equation}
    then as $t\to\infty$, the sequence $\mX_t \to \mU \mV^\top$. To see this, one should plot the univariate cubic function $f(x) \defeq \tfrac{3}{2} \cdot x - \tfrac{1}{2}\cdot x^3$ and see that, for $0 < x < \sqrt{3}$, iterating this cubic will push $x$ closer and closer to $+1$. The final step is to realize that the effect of the iteration in \cref{eq:newton-schulz} is to apply this cubic $f(x)$ to each singular value of $\mX_t$. This also shows that the spectral normalization $\mX_0 = \mG / \norm{\mG}_{\ell_2 \to \ell_2}$ is stronger than what is required: we need only ensure that $\mX_0$ has all singular values greater than zero and less than $\sqrt{3}$ in order for the iteration to converge.

    There are in fact a family of degree $2n+1$ polynomial iterations of the form
    \begin{equation}
    \mX_{t+1} = a \cdot \mX_t + b  \cdot \mX_t \mX_t^\top \mX_t + c \cdot (\mX_t \mX_t^\top)^2 \mX_t + ... + z \cdot (\mX_t \mX_t^\top)^n \mX_t
    \end{equation}
    for suitable $a,b,c,...,z$ that could be used instead of \cref{eq:newton-schulz}. One should choose coefficients $a, b,c,...,z$ so that the univariate polynomial $g(x) = a \cdot x + b \cdot x^3 + c \cdot x^5 + ... + z\cdot x^{2n+1}$ is a suitable approximation to $\sign(x)$. The coefficients can be tuned graphically to achieve the fastest convergence.

    After posting the first version of this paper on arXiv, we found out that the iteration, at least for fixed coefficients, is classical \citep{kovarik1970iterative,bjoerck1971}.
\end{enumerate}
Which of these methods is most useful in practice may depend on factors such as the condition number of the matrix $\mG$ or the nature of the available computational resources.

\clearpage
\section{Proofs}

\renewcommand{\proofname}{\hspace{-0.315em}}

\subsection*{\cref{prop:steepest}: \nameref*{prop:steepest}} \label{proof:steepest}

\begin{proof}
First, let's study the minimization under the change of variables $\Delta \vw = c \cdot \vt$, where $c \geq 0$ encodes the ``magnitude'' and $\vt$ is a unit vector ($\norm{\vt}=1$) encoding the ``direction'':
\begin{align}
    \min_{\Delta \vw \in \R^n} \left[\vg^\top \Delta \vw + \frac{\lambda}{2} \, \norm{\Delta \vw}^2 \right] &= \min_{c \geq 0} \min_{\vt\in\R^n : \norm{\vt}=1} \left[c \cdot \vg^\top \vt + \frac{\lambda}{2} c^2 \norm{\vt}^2 \right] \\
    &= \min_{c \geq 0} \left[c \cdot \min_{\vt\in\R^n : \norm{\vt}=1} \left[\vg^\top \vt\right] + \frac{\lambda}{2} c^2 \right] \label{eq:changed} \\
    &= \min_{c \geq 0} \left[ - c \cdot \norm{\vg}^\dagger+ \frac{\lambda}{2} c^2 \right], \label{eq:changed-again}
\end{align}
Inspecting \cref{eq:changed}, we see that the minimizer for the direction $\vt$ is given by:
\begin{align}
    \vt &= \argmin_{\vt\in\R^n : \norm{\vt}=1} \left[\vg^\top \vt\right] = - \argmax_{\vt\in\R^n : \norm{\vt}=1} \left[\vg^\top \vt\right]
\end{align}
And similarly, by inspecting \cref{eq:changed-again}, the minimizer for the magnitude $c$ is given by:
\begin{align}
    c &= \argmin_{c \geq 0} \left[ - c \cdot \norm{\vg}^\dagger + \frac{\lambda}{2} c^2 \right] = \frac{\norm{\vg}^\dagger}{\lambda}.
\end{align}
Multiplying these expressions, we obtain the minimizer for $\Delta \vw$, yielding the result.
\end{proof}

\subsection*{\cref{prop:sign-descent}: \nameref*{prop:sign-descent}} \label{proof:sign-descent}

\begin{proof}
The result follows by applying \cref{prop:steepest}. We just need that $\argmax_{\norm{\vt}_\infty = 1} \vg^\top \vt = \sign(\vg)$, and also that the dual norm $\norm{\vg}_\infty^\dagger \defeq \max_{\norm{\vt}_\infty = 1} \vg^\top \vt = \vg^\top\sign(\vg)= \norm{\vg}_1$.
\end{proof}

\subsection*{\cref{prop:structural-sign-descent}: \nameref*{prop:structural-sign-descent}} \label{proof:structural-sign-descent}

\begin{proof}
The result follows from \cref{prop:steepest-modular} by setting all the scalars $s_1,...,s_L$ to one and all the norms $\norm{\cdot}_1, ..., \norm{\cdot}_L$ to the $\ell_1$ to $\ell_\infty$ operator norm. All we need is to show that the argmax at each matrix space $l=1,...,L$ satisfies:
\begin{equation}
    \argmax_{\norm{\mT_l}_{\ell_1 \to \ell_\infty} = 1} \trace(\mG_l^\top \mT_l) = \sign(\mG_l).
\end{equation} 
But this holds because, by \cref{prop:tractable-norms}, $\norm{\mT}_{\ell_1 \to \ell_\infty} = \max_i \norm{\mathrm{col}_i(\mT)}_\infty = \max_{ij} \abs{\mT_{ij}}$, and therefore all components in the argmax must be of unit size and gradient aligned.
\end{proof}

\subsection*{\cref{prop:projection}: \nameref*{prop:projection}} \label{proof:projection}

\begin{proof}
To begin, we observe that the minimizer over semi-orthogonal matrices of the ``distance'' $\norm{\mA - \mG}_F$ is the same as the maximizer over semi-orthogonal matrices of the ``alignment'' $\langle \mA, \mG\rangle$, where $\langle\cdot,\cdot\rangle$ denotes the Frobenius inner product. This is because:
\begin{align}
    \norm{\mA - \mG}_F^2 &= \norm{\mA}_F^2 - 2 \cdot \langle \mA, \mG \rangle + \norm{\mG}_F^2,
\end{align}
and the term $\norm{\mA}_F^2$ is fixed at $\norm{\mA}_F^2 = \min(m,n)$ for a semi-orthogonal matrix $\mA \in \mathcal{O}_{m\times n}$.

Now, let $\mG = \sum_i \sigma_i \, \vu_i\vv_i^\top$ denote the SVD of $\mG$. Then the alignment satisfies:
\begin{align}\label{eq:bound}
    \langle \mA, \mG \rangle = \trace \sum_i \sigma_i \,\vv_i\vu_i^\top \mA = \sum_i \sigma_i \, \vu_i^\top \mA \vv_i \leq \sum_i \sigma_i,
\end{align}
where the second equality follows by the cyclic property of the trace, and the inequality is since $\mA$ being semi-orthogonal means that $\vu^\top \mA \vv \leq 1$ for any two unit vectors $\vu$ and $\vv$.

Next, observe that for the semi-orthogonal matrix $\mA_\star = \sum_i \vu_i \vv_i^\top$, we have that:
\begin{equation}
    \langle \mA_\star, \mG \rangle = \sum_i \sigma_i \sum_j \vu_i^\top \vu_j \vv_j^\top \vv_i = \sum_i \sigma_i,
\end{equation}
since the $\{\vu_i\}$ and $\{\vv_i\}$ are orthonormal. Comparing against \cref{eq:bound}, we see that $\mA_\star$ indeed maximizes the alignment, since it achieves the upper bound of $\sum_i \sigma_i$. And $\mA_\star$ therefore also minimizes the distance $\norm{\mA - \mG}_F$ amongst semi-orthogonal matrices $\mA$. Note that if $\mU$ is the matrix that has the $\{\vu_i\}$ as columns, and likewise for $\mV$ and the $\{\vv_i\}$, then this solution may equivalently be expressed as $\mA_\star = \mU\mV^\top$.

All that remains is to explore the uniqueness of this solution:
\begin{itemize}
    \item If $\mG$ is full rank, the solution $\mA_\star$ is unique. $\mG$ being full rank means that all the singular values $\sigma_i$ are positive. In this case, we see from \cref{eq:bound} that to maximize the alignment the semi-orthogonal matrix $\mA$ must satisfy $\vu_i^\top \mA \vv_i =1$ for all $i$. Since $\mA$ has spectral norm one, in turn this requires that
$\mA \vv_i = \vu_i$ and $\mA^\top \vu_i = \vv_i$ for all $i$. These conditions uniquely pick out $\mA = \sum_i \vu_i \vv_i^\top$.
    \item If $\mG$ is not full rank then the solution $\mA_\star$ is not unique. This solution is just as good:
    \begin{equation}
        \mA_\dagger = \sum_{i:\sigma_i > 0} \vu_i \vv_i^\top + \sum_{i:\sigma_i = 0}\vu_i (-\vv_i)^\top.
    \end{equation}
\end{itemize}
This completes the proof.\end{proof}

\subsection*{\cref{prop:shampoo-steepest}: \nameref*{prop:shampoo-steepest}} \label{proof:shampoo-steepest}

\begin{proof}
First, we apply \cref{prop:steepest-modular} with scalars $s_1,...,s_L$ set to one and all norms set to $\norm{\cdot}_{\ell_2\to\ell_2}$. This tells us that the solution is given by $\Delta \mW_l = - \eta\cdot \argmax_{\norm{\mT_l}_l=1} \trace(\mG_l^\top \mT_l)$ for each $l=1,...,L$ and with $\eta = \frac{1}{\lambda}\sum_{k=1}^L \norm{\mG_k}_{\ell_2\to\ell_2}^\dagger$. We just need to resolve the dual norm and evaluate the argmax.

Let's start with the dual norm. For a matrix $\mG$ with SVD $\sum_i \sigma_i \, \vu_i \vv_i^\top = \mU \mSigma \mV^\top$ we have:
\begin{align}
    \norm{\mG}_{\ell_2\to\ell_2}^\dagger \defeq \max_{\norm{\mT}_{\ell_2\to\ell_2}=1}\trace \mG^\top \mT &= \max_{\norm{\mT}_{\ell_2\to\ell_2}=1} \trace \sum_i \sigma_i \, \vv_i \vu_i^\top \mT \\ &= \max_{\norm{\mT}_{\ell_2\to\ell_2}=1}\sum_i \sigma_i\, \vu_i^\top \mT \vv_i \leq \sum_i \sigma_i = \trace \mSigma,
\end{align}
where the upper bound follows from the spectral norm constraint on $\mT$. But this upper bound is attained by setting $\mT = \mU\mV^\top$ (also resolving the argmax) and so $\norm{\mG}_{\ell_2\to\ell_2}^\dagger = \trace \mSigma$.

The uniqueness claim follows by the same argument as for \cref{prop:projection}. \end{proof}

\subsection*{\cref{prop:majorization}: \nameref*{prop:majorization}} \label{proof:majorization}

\begin{proof}
    First observe that the square loss is quadratic in $\mW$ so there are no cubic terms or higher. The bound must agree to first-order with the first-order Taylor expansion of $\el(\mW+\Delta \mW)$, which is precisely $\el(\mW) + \langle\nabla_\mW \el(\mW), \Delta \mW\rangle$, since otherwise the bound would be violated for sufficiently small $\Delta\mW$. To obtain the second-order piece of the bound, it's easiest just to multiply out $\el(\mW+\Delta\mW)$ and see that the second-order piece of $\el(\mW+\Delta \mW)$ satisfies:
    \begin{align}
        \frac{1}{2n}\sum_{i=1}^n\frac{1}{d_\out}\norm{\Delta \mW \vx^{(i)}}_2^2 \leq \frac{1}{2n}\sum_{i=1}^n\frac{1}{d_\out}\norm{\Delta \mW}_{\ell_2\to\ell_2}^2 \cdot \norm{\vx^{(i)}}_2^2
        =\frac{1}{2}\frac{d_\inn}{d_\out}\norm{\Delta \mW}_{\ell_2\to\ell_2}^2,
    \end{align}
    where the last equality uses the input normalization $\norm{\vx^{(i)}}_2 = \sqrt{d_\inn}$. We are done.
\end{proof}

\subsection*{\cref{prop:steepest-modular}: \nameref*{prop:steepest-modular}} \label{proof:steepest-modular}

\begin{proof}
For each layer $l=1,...,L$, we decompose $\Delta\mW_l$ into its magnitude and direction: $\Delta \mW_l = c_l \cdot \mT_l$, for $c_l \geq 0$ and $\norm{\mT_l}_l = 1$. Under this change of variables, the minimization becomes:
\begin{align}
    &\min_{\Delta \mW_1, \dots, \Delta \mW_L} \left[ \sum_{l=1}^L \langle \mG_l, \Delta \mW_l \rangle + \frac{\lambda}{2} \max_{l=1}^L s_l^2 \norm{\Delta\mW_l}_l^2 \right] \\
    &\qquad= \min_{c_1, \dots, c_L \geq 0} \left[ \sum_{l=1}^L c_l \min_{\norm{\mT_l}_l=1} \langle \mG_l, \mT_l \rangle + \frac{\lambda}{2} \max_{l=1}^L s_l^2 c_l^2 \right] \label{eq:many-var}\\
    &\qquad= \min_{c_1, \dots, c_L \geq 0} \left[ -\sum_{l=1}^L c_l \norm{\mG_l}_l^\dagger + \frac{\lambda}{2} \max_{l=1}^L s_l^2 c_l^2 \right]\label{eq:many-var2}\\
    &\qquad= \min_{\eta\geq 0} \left[ -\sum_{l=1}^L \frac{\eta}{s_l} \norm{\mG_l}_l^\dagger + \frac{\lambda}{2} \eta^2 \right], \label{eq:one-variable}
\end{align}
where \cref{eq:one-variable} follows by observing that at the minimum we must have $s_1 c_1, ..., s_L c_L$ all taking the same value of $\eta\geq 0$ (still to be determined), since otherwise we could increase the sum $\sum_l c_l \norm{\mG_l}_l^\dagger$ by increasing any of the slack $c_l$ without paying a penalty in terms of the max. We can now read off the minimizers from \cref{eq:many-var,eq:many-var2,eq:one-variable}:
\begin{align}
    \mT_l &= \argmin_{\norm{\mT_l}_l = 1} \, \langle \mG_l, \mT_l \rangle = - \argmax_{\norm{\mT_l}_l = 1} \, \langle \mG_l, \mT_l \rangle; \\
    c_l &= \frac{\eta}{s_l}; \\
    \eta &= \frac{1}{\lambda} \sum_{k=1}^L \frac{1}{s_k} \norm{\mG_k}_k^\dagger. \label{eq:eta}
\end{align}
Combining, we obtain the overall minimizer for each $l=1,...,L$ via $\Delta \mW_l = c_l \cdot \mT_l = - \frac{\eta}{s_l} \argmax \, \langle \mG_l, \mT_l \rangle$, where $\eta$ is given by \cref{eq:eta}, proving the result.
\end{proof}

\subsection*{\cref{prop:tractable-norms}: \nameref*{prop:tractable-norms}} \label{proof:tractable-norms}

\begin{proof}
Let's start with the $\ell_1\to\ell_p$ operator norm. Here we observe that, in matrix-vector multiplication, each component of an input vector selects and scales a column of the matrix:
\begin{align}
    \norm{\mM}_{\ell_1 \to \ell_p} = \max_{\norm{\vx}_1 = 1} \norm{\mM \vx}_p = \max_{\norm{\vx}_1 = 1} \Big\|\sum_j \mathrm{col}_j(\mM) \vx_j\Big\|_p &\leq \max_{\norm{\vx}_1 = 1} \sum_j \abs{\vx_j}\cdot \norm{\mathrm{col}_j(\mM)}_p\\
    &\leq \max_{\norm{\vx}_1 = 1} \norm{\vx}_1 \cdot \max_j \norm{\mathrm{col}_j(\mM)}_p\\
    &= \max_j \norm{\mathrm{col}_j(\mM)}_p,\label{eq:upper-bound}
\end{align}
by the triangle inequality and H\"older's inequality. But the upper bound in \cref{eq:upper-bound} is attained by selecting the column index $j_\star = \argmax_j \norm{\mathrm{col}_j(\mM)}_p$ with the largest norm, then setting $\vx_{j_\star}=1$ and the other input components to zero. So $\norm{\mM}_{\ell_1 \to \ell_p} = \max_j \norm{\mathrm{col}_j(\mM)}_p.$

Next, let's deal with the $\ell_p \to \ell_\infty$ operator norm. Here we break up a matrix-vector product in terms of the dot product between the vector and the matrix rows:
\begin{align}
    \norm{\mM}_{\ell_p \to \ell_\infty} = \max_{\norm{\vx}_p = 1} \norm{\mM \vx}_\infty &= \max_{\norm{\vx}_p = 1} \max_i \abs{\vx^\top \mathrm{row}_i(\mM) } \\
    &= \max_i \max_{\norm{\vx}_p = 1} \abs{\vx^\top \mathrm{row}_i(\mM) } \\
    &= \max_i \norm{\mathrm{row}_i(\mM)}_p^\dagger.
\end{align}
The proof is completed by recalling that the vector $\ell_p$ norm is dual to the vector $\ell_q$ norm for $1/p + 1/q =1$. In other words, $\norm{\cdot}_p^\dagger = \norm{\cdot}_{\frac{p}{p-1}}$.
\end{proof}

\end{document}